\documentclass[runningheads]{llncs}
\usepackage{graphicx}
\usepackage{caption}
\usepackage{graphicx}
\usepackage{float} 
\usepackage{subcaption}
\usepackage{amsfonts,amssymb}
\usepackage{booktabs}
\usepackage{multirow}
\usepackage[table,xcdraw]{xcolor}
\usepackage{bigstrut}
\usepackage{booktabs}

\usepackage[linesnumbered,lined,boxed,commentsnumbered,ruled,vlined]{algorithm2e}
\begin{document}
\title{Improving CTC-AED model with integrated-CTC and auxiliary loss regularization}
\author{Daobin Zhu\inst{1}\and
Xiangdong Su\inst{1} \and
Hongbin Zhang\inst{2}
}
\authorrunning{Zhu. et al.}
\titlerunning{Improved CTC-AED model with integrated-CTC and auxiliary loss reg.}
\institute{Inner Mongolia University, Huhhot, China \and Xinjiang University, Urumuqi, China
	\\
}
\maketitle              % typeset the header of the contribution
\begin{abstract}
Connectionist temporal classification (CTC) and attention-based encoder decoder (AED) joint training has been widely applied in automatic speech recognition (ASR). Unlike most hybrid models that separately calculate the CTC and AED losses, our proposed integrated-CTC utilizes the attention mechanism of AED to guide the output of CTC. In this paper, we employ two fusion methods, namely direct addition of logits (DAL) and preserving the maximum probability (PMP). We achieve dimensional consistency by adaptively affine transforming the attention results to match the dimensions of CTC. To accelerate model convergence and improve accuracy, we introduce auxiliary loss regularization for accelerated convergence. Experimental results demonstrate that the DAL method performs better in attention rescoring, while the PMP method excels in CTC prefix beam search and greedy search.

\keywords{Speech recognition  \and Hybrid CTC and attention \and Two-pass.}
\end{abstract}
\section{Introduction}
Due to its outstanding recognition performance, end-to-end (E2E) speech recognition has been increasingly applied in both academic and industrial fields. There are three mainstream E2E models in ASR, namely CTC-based \cite{lee2021intermediate,graves2006connectionist}, Transducer-based \cite{graves2012sequence,li2021better,graves2013speech}, and AED-based  models\cite{chiu2017monotonic,kim2017joint,chorowski2015attention,chan2016listen,tian2020synchronous}. Based on whether the decoding stage considers the historical information of frame-wise outputs, they can be further classified into autoregressive (AR) \cite{chan2020imputer} and non-autoregressive (NAR) \cite{komatsu2022non,higuchi2020mask,ghazvininejad2019mask,fujita2020insertion} models. AR models typically have higher accuracy but longer decoding time, while NAR models exhibit relatively poorer recognition performance but faster decoding speed.

The CTC-based models have a decent decoding speed. However, due to their highly unreasonable assumption of context independence and the lack of language modeling capabilities, these models fail to meet practical requirements. Recently, a hybrid CTC-AED model \cite{zhang2020unified} has been proposed. This model combines CTC and AED losses, applies dynamic chunk attention, and performs two-pass decoding. The second-pass decoding by AED (AR pattern) significantly improves the accuracy of the first-pass decoding by CTC (NAR pattern). However, in the training stage, the CTC and AED loss functions are separately computed, and the only relationship between CTC and AED is a weighted sum during the calculation of the total loss. To address this limitation, we propose a structure called integrated-CTC, which fuses the results from the attention-based decoder into CTC during the training stage. This approach helps alleviate the inherent weakness of CTC in language modeling to some extent. The AR method (attention mechanism) provides more contextual information to CTC and helps the encoder form richer modeling capabilities. By fusing the results of the AR and NAR methods during the training stage, we achieve competitive character error rate (CER) in the decoding stage without the need for a rescoring process. Furthermore, we found that using the direct addition of logits (DAL) method reduced the character error rate relatively (CERR) by 1\% in the results obtained solely from the one-pass decoding, compared to the corresponding decoding approach in WeNet \cite{yao2021wenet}. The method of preserving the maximum probability (PMP) achieves a CER of 4.79\% through the ctc prefix beam search decoding, outperforming the result of 4.93\% obtained by the same method in WeNet.

The essence of two-pass decoding is to utilize the first pass for rapid decoding and improve accuracy in the second pass. Currently, mainstream two-pass hybrid models do not use a higher-accuracy AR model to correct the output of CTC during the training stage. Our proposed integrated-CTC allows CTC to reference the results from the attention-based decoder during output generation in the training stage. We achieve dimensional consistency between the attention-based decoder results and CTC outputs by employing the proposed adaptive affine algorithm to scale the dimensions of the AED results to match the dimensions of CTC. Through frame-level correction, the output of CTC is regularized, resulting in fairly good recognition results. The total loss of integrated-CTC is obtained by weighting the integrated-CTC loss and the attention loss. The specific experimental details regarding the impact of loss on CER will be shown in Table \ref{tab:table1}.

Some recent work has focused on the regularization of the CTC loss, aiming to achieve better convergence by modifying the number of CTC layers or the structure of CTC \cite{lee2021intermediate,komatsu2022non,2021Relaxing}. This regularization of the CTC loss not only avoids significant computational overhead but also significantly reduces the difficulty of model training. To a certain extent, this represents a significant improvement for CTC, as it reduces computational costs and improves accuracy. However, relying solely on the CTC structure for overall improvement has limitations, and in \cite{lee2021intermediate}, despite the use of a conformer model in the encoder, the CER for AISHELL-1 only reached 5.2\%. This is also why we chose the CTC-AED model. Based on the CTC-AED model, we also propose auxiliary loss regularization to help the model achieve better recognition performance and faster convergence speed. Our experimental results show that using the integrated-CTC loss reduces training time by 5\% compared to using the official loss in WeNet.

The main contributions of this paper are as follows:

1. We propose a simple yet effective training method called integrated-CTC, which influences the output of CTC through the attention mechanism during the training stage.

2. We introduce an algorithm called adaptive affine that dynamically adjusts the dimensions of AED outputs.

3. We propose auxiliary loss regularization to facilitate faster convergence and improve model accuracy.

4. We demonstrate the impact of assigning different weights to posterior probabilities in attention and achieve a CER of 4.49\% on the AISHELL-1 dataset using two-pass decoding. Furthermore, by using only greedy search decoding, we achieve a CER of 4.79\%.
\begin{figure*}[]
	\centering
	\includegraphics[width=1.0\linewidth]{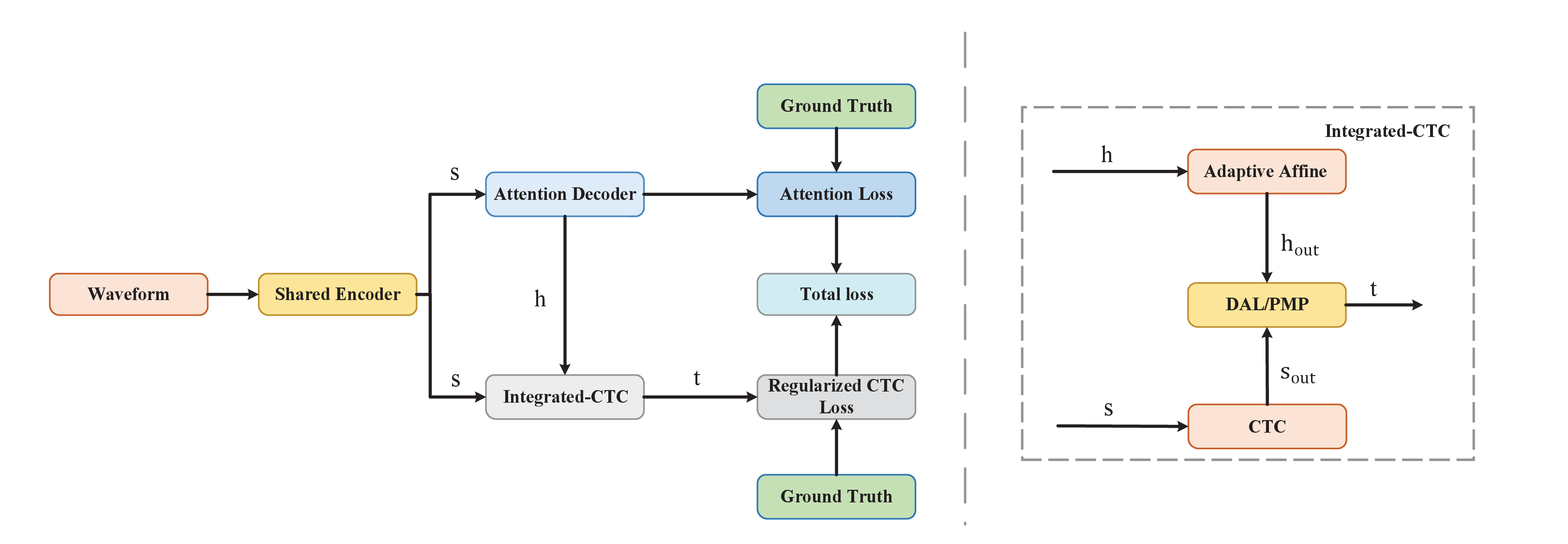}
	%\subcaption{The architecture of CTC-AED hybrid model in two-pass approch}
	\caption{\textbf{\textit{(Left)}} shows the structure of our proposed model. The model consists of three parts: shared encoder, attention decoder, and integrated-CTC. Unlike other hybrid models, we integrate the output of the attention decoder to the CTC to help the CTC to decode accurately.
		\textbf{\textit{(Right)}} shows the internal construction of integrated-CTC. It consists of three parts: adaptive affine, CTC and integration algorithm. Among them, adaptive affine is responsible for aligning the output dimension of the attention decoder with the CTC and corresponding between frames. The posterior probability vectors of attention are integrated with the output of CTC by DAL or PMP algorithm after adaptive affine.}
	\label{figure:1}%文中引用该图片代号
\end{figure*}

\section{Hybrid CTC-AED Model with Two-pass Decoding}
\subsection{Model Architecture} 
The proposed hybrid CTC-AED model \cite{kim2017joint} with two-pass \cite{sainath2019two,sainath2020streaming} decoding is illustrated in Fig. \ref{figure:1} \textbf{\textit{(Left)}}. The speech input is fed into a shared encoder, attention decoder, and integrated-CTC module. The total loss is computed by combining the integrated-CTC loss and attention loss. The shared encoder produces the vector $s$, which is then processed by the attention decoder to obtain $h$. The integrated-CTC module, as shown in Fig. \ref{figure:1} \textbf{\textit{(Right)}}, involves obtaining $s_{out}$ through CTC and transforming $h$ into $h_{out}$ using the adaptive affine algorithm. Finally, $s_{out}$ and $h_{out}$ are fused using two fusion methods.

\subsubsection{Shared Encoder of the Hybrid Model}
The shared encoder in the hybrid model consists of multiple encoder blocks, responsible for transforming speech features into feature vectors. The encoder can be based on Transformer or Conformer architecture, and in our experiments, we utilize Conformer \cite{2020Conformer}. The Conformer network primarily comprises of relative positional multi-head attention modules, position-wise feed forward modules, and convolution modules. The relative positional encoding in multi-head attention provides robustness and generalizability to inputs of different lengths. The position-wise feed forward network linearly maps inputs at each time step and changes the dimensions of input vectors through the average forward matrix. The convolution module employs causal convolution to ensure that the model does not rely on right-context information. The computation formula for a Conformer block is as follows:

\begin{equation}
	x_i'=x_i+\frac{1}{2}\mathbf{FFN}(x_i),
\end{equation}

\begin{equation}
	x_i''=x_i'+\mathbf{MHA}(x_i'),
\end{equation}
\begin{equation}
	\hat{x}=x_i''+\mathbf{Conv}(x_i''),
\end{equation}
\begin{equation}
	y_i=\mathbf{Batchnorm}(\hat{x}+\frac{1}{2}\mathbf{FFN}(\hat{x})),
\end{equation}
where $x_i\in \mathbb{R}^{B\times L \times D}$, B, L and D indicate batch size, sequence length and dimension of speech features respectively. Feed forward network (FFN) contains two linear layers and a swish activation function. $x_i'$ is successively entered into three modules for calculation and residual connection. After a final layer of FFN the output is fed to a Batchnorm function.

%Where $x_i$ has dimensions of B x L x D, where $B$, $L$, and $D$ represent batch size, sequence length, and speech feature dimensions, respectively. The forward neural network consists of two linear layers and a swish activation function. $X_{i'}$ is calculated by passing through three consecutive modules and is then connected with the residual connection. The final layer of the feed forward network is fed into the BatchNorm function.
\subsubsection{Two-pass Decoder And Auxiliary Loss Regularization}
The output of the shared encoder yields an output $\mathbf{O}$ with dimension $T\times V$, where T represents the number of frames in the audio and V represents the size of the vocabulary. CTC finds N best paths (e.g., using beam search) with the highest probabilities based on the corresponding decoding algorithm. The attention decoder performs the second-pass decoding. The output $\mathbf{O}$ is fed into the attention decoder, which performs rescoring in an AR manner. The conditional probability $P_{AR}(Y|O)$ can be represented as:
\begin{equation}
	P_{AR}(Y|O)=P(y_1|O)\sum^L_{i=2}P(y_i|y_{<i},O),
\end{equation}
where $y_{i}$ represents the i-th token of the predicted sequence with length L. The attention-based decoder uses a Transformer network structure. During the training process, both CTC and the attention-based decoder compute their respective losses, $L_{CTC'}$ and $L_{AR}$, which are then combined using a weighted sum operation. $L_{CTC'}$ represents the loss after fusion of the output vectors from CTC and the attention decoder, while $L_{CTC}$ is the loss from CTC without fusion. Additionally, we also tried adding $L_{CTC}$ to the total loss, but found that it resulted in worse experimental performance. We speculate that too many CTC regularization terms may lead to overfitting, disregarding the intrinsic information in the speech data itself. The loss function can be expressed as follows:
\begin{equation}
	\mathcal{L}=\alpha\mathcal{L}_{CTC'}+(1-\alpha)\mathcal{L}_{AR},
\end{equation}
where $\alpha$ is a hyperparameter. In the loss computation, we tested different values of $\alpha$ ranging from 0.1 to 0.9, and found that 0.5 provided the most stable results. We employed two fusion algorithms to combine the output from the attention-based decoder with the output from CTC.

The dimension of $ t $ is $[L, D]$, where L represents the length of the audio and D represents the size of the dictionary. Adding the probabilities of $y_{AED}$ to $y_{CTC}$ can help improve the recognition performance of CTC. Additionally, we designed an algorithm named PMP that saves the maximum probability for each frame and sets the probabilities of other positions to 0, aiming to reduce the number of operations. The experimental details will be presented in Section \ref{ssec:results}.

\subsubsection{Regularized CTC}
As mentioned earlier, in most CTC-AED hybrid models (e.g., ESPNet and WeNet), the CTC loss and AED loss are computed separately during training. After passing through the shared encoder, the outputs are decoded by the CTC decoder and AED decoder, and their respective loss functions are weighted and summed to obtain the total loss for updating the model. However, the attention mechanism can help CTC obtain more accurate results, as demonstrated by the need for attention-based second-pass decoding to correct certain erroneous CTC outputs. However, the AR model increases decoding time. Delay rate is also important in speech recognition. Let's consider an alternative approach. If we use the attention mechanism to guide CTC decoding during training, we can achieve competitive recognition results without the need for second-pass decoding. This means that even with one-pass decoding (e.g., CTC beam search), we can obtain similar recognition results as with second-pass decoding. In the training process, we employ the same multi-objective learning framework as in \cite{watanabe2018espnet} to improve the accuracy and robustness of the model. Specifically, we combine the fusion CTC loss $L_{CTC'}$ and the attention-based cross-entropy loss $L_{AR}$. Furthermore, we use the adaptive affine algorithm to match the dimension of the attention-based decoder output $y_{AED}$ with the dimension of the CTC decoder output $y_{CTC}$. Here, $y_{AED}$ is a vector of length L with a dimension of $V$, while $y_{CTC}$ is a vector of length $P$ with a dimension of $V$. In practice, $P$ is often longer than $L$, but the length of the speech sequence is not fixed, so we cannot set the dimension of the linear mapping as a fixed value. Therefore, we save the maximum probability or the complete posterior probability of each frame in $y_{AED}$. By expanding the dimension of $y_{AED}$ to match the dimension of $y_{CTC}$, mapping $L$ to $P$, the final posterior probability of $y_{CTC}$ is obtained by weighted summation of $y_{CTC}$ and $y_{AED_{L->P}}$. The expression for $y_{CTC}$ is as follows:
\begin{equation}
	y_{CTC}\colon=\lambda y_{AED_{L\to P}}+y_{CTC},
\end{equation}
where $\lambda$ represents the hyperparameter for the weight of $y_{AED}$. The detailed experimental details for this part will be presented in Table \ref{tab:table1} .

\subsection{Adaptive Affine Transformation}
The attention-based decoder and CTC decoder produce outputs of different dimensions, making direct fusion challenging. To address this issue, we propose an adaptive affine transformation algorithm to dynamically adjust the output dimension of the attention-based decoder to match that of the CTC decoder. This algorithm prepares for the subsequent step of using the attention-based decoder to correct the output of the CTC decoder. The output of the attention-based decoder corresponds to the output of the CTC decoder and is linearly combined with it. Algorithm \ref{algorithm} provides the pseudocode for the adaptive affine transformation algorithm.

For the output of the CTC decoder, multiple consecutive frames may correspond to a single phoneme, while for the output of the attention-based decoder, one frame corresponds to one phoneme. Therefore, the output of the attention-based decoder needs to be extended to match the dimension of the CTC decoder's output. We uniformly expand the logits of the attention-based decoder to ensure that most frames correspond to the frames of the CTC decoder. In fact, although there are some frames in the output of the attention-based decoder ($y_{AED}$) that do not have a one-to-one correspondence with the output of the CTC decoder ($y_{CTC}$), they do not have a significant impact on the overall experimental results. The effect of the frames in the output of the attention-based decoder not strictly aligning with the frames of the CTC decoder can be alleviated by assigning weights to the dimension-matched vectors.
%Algorithm 1
\def\SetClass{article}

\IncMargin{1em}
\begin{algorithm}
	\SetKwData{totallength}{total_length}
	\SetKwFunction{torchcat}{torch.cat}\SetKwFunction{size}{size}\SetKwFunction{torchzeros}{torch.zeros}
	\SetKwFunction{INT}{int}\SetKwFunction{unsqueeze}{unsqueeze}\SetKwFunction{expand}{expand}
	
	\# $y\_h$, $a\_o$: the output of CTC and attention decoder respectively\\
	\# $a\_o'$: updated $a\_o$\\
	$total\_length$ = 0 \\
	$repeat\_time$ =  \INT($y\_h$.\size()[1] / $a\_o$.\size()[1]) + 1\\
	$a\_o'$ = \torchzeros(1, $y\_h$.\size()[1], $y\_h$.\size()[2])\\
	\For{$i \leftarrow 1$ \KwTo a\_o.\size$()[0]$}{
		$tmp$ = $a\_o$[i][0].\unsqueeze(0) \\
		\For{$j \leftarrow 1$ \KwTo a\_o.\size$()[1]$}{
			$repeate$ = $a\_o$[i][j].\unsqueeze(0) \\
			\lIf{$total\_length$ \textless $y\_h$.\size$()[1]$}{}
			\qquad$repeat$ = $repeat$.\expand($repeat\_time$, $a\_o$.\size()[2])\\
			\qquad$tmp$ = \torchcat(($tmp$, $repeat$), dim=0) \\
			\qquad$total\_length$ $=$ $repeat + total\_length$}
		$tmp$ = $tmp$[1:][:]\\
		\lIf{$total\_length$ $!=$ $y\_h$.\size $()[1]$}{}
		\qquad $tmp$ = $tmp[:][$:-($total\_length$-$y\_h$.\size()[1])].\unsqueeze(0)\\
		\qquad $a\_o'$ = \torchcat(($a\_o'$, $tmp$), dim=0)\\
		\lElse{}{
			\qquad $tmp$ = $tmp$.\unsqueeze(0)\\
			\qquad $a\_o'$ = \torchcat(($a\_o'$, $tmp$), dim=0)}\\}	
	\caption{Adaptive affine in a PyTorch-like style}\label{algo_disjdecomp}
	\label{algorithm}
\end{algorithm}\DecMargin{1em}

\section{Experiments}
\label{sec:Experiments and Results}

\subsection{Dataset}
\label{sec:Dataset}

In our work, all experiments were conducted on AISHELL-1, a dataset recorded by AISHELL and spoken by 400 people for a total of 178 hours. The training set was recorded by 340 people, the test set by 20 people and the validation set by 40 people, each recording an average of 300 sentences. The words spoken by each person are placed in a folder.
\subsection{Experimental Setup}
\label{sec:Experimental conditions}
For all our experiments, we use 80-dimensional log-mel filter bank (FBank) features computed on a 25ms window with a 10ms shift. The modelling unit consists of 4233 characters, of which 4230 are Chinese characters, a $<$blank$>$ character for CTC output, a $<$unk$>$ character for unknown characters, and a $<$sos/eos$>$ for start-of-sentense and end-of-sentence.We also do speed perturb with 0.9, 1.0 and 1.1 on the whole data. SpecAugment \cite{park2019specaugment} with 2 frequency masks where the maximum frequency mask F = 10, and 2 time masks where the maximum time mask T = 50.

The hybrid model encoder consists of 12 conformer blocks, and the decoder consists of 6 transformer blocks. The multi-head attention has 4 heads. The subsampling network in the convolutional module includes 2D convolutional layers (subsampled by a factor of 4 in frame rate) with a Swish activation function. The kernel size is set to 3, stride is 2, and the number of channels is 256. The hidden units size of the feed forward neural layer is set to 2048. We used the Adam optimizer with a warm-up of 25,000 steps, and the learning rate scheduling described in \cite{vaswani2017attention} to train the models. Additionally, after 210 epochs, we selected the top 30 models based on their loss values and performed parameter averaging to export the best model. To evaluate the inference speed, we measured the real-time factor (RTF) on the test set. The reported RTF results are based on the test set, not the validation set.

\section{Results}
\label{ssec:results}
\subsection{Comparison of Different Decoding Coefficients}
In this section, we compare different values of the parameter $\lambda \in (0.01, 0.09)$ for the two fusion algorithms. The weight $\lambda$ represents the degree to which the attention-based decoder influences the output of the CTC decoder. A higher value of $\lambda$ corresponds to a greater influence of the attention-based decoder. As shown in Table \ref{tab:table1}, it is evident that the optimal value for $\lambda$ is 0.05. Since the results from the attention-based decoder are generated first during training, it provides prior knowledge to the CTC model, and the weight adjustment determines the extent of its influence on the CTC model. When $\lambda$ ranges from 0.03 to 0.07, the model achieves better recognition performance. Although using only one-pass decoding yields good results, the best performance is still achieved by finding the N best CTC paths and rescoring them with the attention-based decoder. By comparing the two fusion algorithms, it is apparent that the fusion method of DAL from $y_{AED}$ to $y_{CTC}$ is more effective for the attention-based decoding approach. Furthermore, the algorithm that PMP helps the CTC-based decoding approach achieve better recognition performance. We speculate that retaining only the highest probability results in the probability distribution allows CTC to quickly identify high-scoring paths, effectively performing pruning operations. This is particularly beneficial for the one-pass decoding method.

\setlength{\tabcolsep}{0.5mm}{

	% \usepackage{multirow}
	\begin{table}[htbp]
		\centering
		\caption{Comparision of the two integration algorithms with weight $\lambda$.(CER \%)}
		\label{tab:table1}
		\begin{tabular}{clccccc}
			\toprule
			\multicolumn{2}{c}{Weight $\lambda$}              & \multicolumn{1}{l}{0.01} & \multicolumn{1}{r}{0.03} & 0.05          & 0.07 & 0.09 \\ \hline
			\multirow{4}{*}{DAL}  & attention decoder      & 5.23                     & 4.89                     & \textbf{4.85} & 4.96 & 5.03 \\
			& attention rescore      & 4.76                     & 4.52                     & \textbf{4.49} & 4.55 & 4.56 \\
			& ctc gready search      & 5.20                     & \textbf{4.84}            & \textbf{4.84} & 4.93 & 5.09 \\
			& ctc prefix beam search & 5.20                     & \textbf{4.83}            & 4.84          & 4.93 & 5.09 \\ 
			\midrule
			\multirow{4}{*}{PMP} & attention decoder      & 5.12                     & \textbf{4.87}            & 4.92          & 4.91 & 4.87 \\
			& attention rescore      & 4.78                     & 4.54                     & \textbf{4.50} & 4.56 & 4.58 \\
			& ctc gready search      & 5.11                     & 4.76                     & \textbf{4.79} & 4.92 & 4.92 \\
			& ctc prefix beam search & 5.11                     & 4.76                     & \textbf{4.79} & 4.91 & 4.93 \\ 
			\bottomrule
		\end{tabular}
	\end{table}

\subsection{Comparison of different models}

While studying the parameter influences on the models, we also enhanced the models to achieve better accuracy. Both models used 12 layers of encoders and 6 layers of decoders, and we maintained a consistent testing environment. All the models we compared were attention-based.

All our experiments were conducted on the open-source toolkit WeNet \cite{yao2021wenet}. On the AISHELL-1 dataset, WeNet achieved a CER of 4.64\% using the Conformer as the encoder and attention rescoring. Under the condition of consistent experimental parameters, our model achieved a CER of 4.49\%.  

\begin{table}[htbp]
	\centering
	\caption{Comparision with other conventional, hybrid models (CER\%). All models in this table use SpecAugment to improve
		the performance.}
	\begin{tabular}{llll} 
		\toprule
		Model          & LM  & Dev & Test  \\ 
		\midrule
		Autoregressive Transformer \cite{2021Relaxing} & w/o & 4.9 & 5.4   \\
		ESPNet \cite{watanabe2018espnet}         & w/o & 4.6 & 5.1  \\
		Autoregressive Conformer \cite{pengchengESPnet}   & w/o & 4.4 & 4.7   \\
		ESPNet2        & w/o & 4.4 & 4.7  \\
		SRU++ \cite{pan2022sru}         & w   & 4.4 & 4.7   \\
		WeNet \cite{yao2021wenet}          & w/o &  -   & 4.6  \\
		\midrule
		integrated-CTC with add way           & w/o & \textbf{4.2} & \textbf{4.5}   \\
		\bottomrule
	\end{tabular}
\end{table}

Our proposed model achieved highly competitive recognition results compared to other hybrid models. Even with only one-pass decoding, our model produced a CER that was nearly identical to the two-pass decoding, demonstrating the effectiveness of our proposed algorithm.

\subsection{Comparison of Decoding Latency}
We built libtorch using C++ and exported the models corresponding to the two fusion algorithms, which were then quantized. The value of parameter $\lambda$ for all models was set to 0.05 to ensure fairness and accuracy in the experiments. We presented the decoding and rescoring latency of different models and tested their RTF. The Runtime CER was measured on the test dataset. Directly summing the posterior probability distributions $y_{AED}$ from AED and $y_{CTC}$ from CTC achieved a lower CER because it is equivalent to providing soft labels for $y_{CTC}$, which contains richer information. By only saving the maximum value of each frame from $y_{AED}$ and adding it to $y_{CTC}$, the decoding latency was relatively reduced. From Table 3, we can also see that if rescoring is not used, the RTF of the model is improved by approximately 1/4, but the CER is increased by 0.3\%. Quantizing the model reduced the RTF by approximately 0.01, while the CER decreased by approximately 0.15\%.
\begin{table}[htbp]
	\caption{Comparison of the decoding time and rescoring time of the two models}
	\label{tab:table3}
	\centering
	\scalebox{0.9}{
		\begin{tabular}{ccccc}
			\toprule
			exp.                    & decode latency & rescore latency & RTF     & CER    \\ \midrule
			Add orig.(float 32)     & 386ms          & 88ms            & 0.0863  & 4.49\% \\
			Add quant.(int 8)       & 337ms          & 77ms            & 0.0753 & 4.56\% \\
			Maximal orig.(float 32) & 371ms          & 86ms            & 0.0829 & 4.50\% \\
			Maximal quant.(int 8)   & 316ms          & 71ms            & 0.0706 & 4.63\% \\ 
			
			\bottomrule
		\end{tabular}
	}
\end{table}
\section{Conclusion}
We found that during model training, the attention-based decoder can positively influence CTC. The fusion of the output from the attention-based decoder and the prediction from CTC helps the model achieve better recognition results. Both fusion methods we proposed effectively improve the model's recognition performance. To ensure the fusion of CTC and the attention-based decoder, we proposed a simple but effective algorithm called adaptive affine. The algorithm of PMP consumes less time for decoding, while DAL can yield more accurate recognition results. Additionally, we compared the CER for four decoding methods, and the experimental results demonstrated that our proposed model achieves more competitive performance. In the future, we plan to expand the proposed methods to larger-scale datasets such as WenetSpeech and GigaSpeech.


\end{document}